\documentclass[conference]{IEEEtran}
\IEEEoverridecommandlockouts
\usepackage{cite}
\usepackage{url}
\usepackage{svg}
\usepackage{comment}
\usepackage{amsmath,amssymb,amsfonts, amsthm}
\usepackage{mathtools}

\usepackage{bm}
\usepackage{mathrsfs}
\usepackage{siunitx}
\usepackage{algorithm}
\usepackage{algorithmic}
\usepackage[acronym]{glossaries}
\usepackage{graphicx}
\usepackage{textcomp}
\usepackage{xcolor}
\usepackage{subcaption}
\usepackage{placeins}
\usepackage{booktabs}
\usepackage[hidelinks]{hyperref}
\def\BibTeX{{\rm B\kern-.05em{\sc i\kern-.025em b}\kern-.08em
    T\kern-.1667em\lower.7ex\hbox{E}\kern-.125emX}}

\DeclareMathAlphabet{\mymathbb}{U}{BOONDOX-ds}{m}{n}

\newacronym{mpc}{MPC}{Model Predictive Control}
\newacronym{deepc}{DPC}{Data-Driven Predictive Control}
\newacronym{narx}{NARX}{Nonlinear Autoregressive Exogenous}

\newtheorem{definition}{Definition}
\newtheorem{lemma}{Lemma}

\usepackage{fancyhdr}
\begin{document}

\title{Less is More: Contextual Sampling for Nonlinear Data-Driven Predictive Control}

\author{Julius Beerwerth$^1$ and Bassam Alrifaee$^1$%
\thanks{Supplementary material is available at:
\href{https://bassamlab.github.io/DeePC-Contextual-Sampling/}{https://bassamlab.github.io/DeePC-Contextual-Sampling/}.}%
\thanks{$^{1}$ The authors are with the Department of Aerospace Engineering, University of the Bundeswehr Munich, Germany, \texttt{firstname.lastname@unibw.de}.}}

\maketitle
\thispagestyle{fancy}

\begin{abstract}
\acrfull{deepc} optimizes system behavior directly from measured trajectories without requiring an explicit model. However, its computational cost scales with dataset size, limiting real-time applicability to nonlinear robotic systems.

For robotic tasks such as trajectory tracking and motion planning, real-time feasibility and numerical robustness are essential. Nonlinear \acrshort{deepc} often relies on large datasets or learned nonlinear representations to ensure accuracy, both of which increase computational demand.
We propose \emph{Contextual Sampling}, a dynamic data selection strategy that adaptively selects the most relevant trajectories based on the current state and reference. By reducing dataset size while preserving representativeness, it improves computational efficiency.

Experiments on a scaled autonomous vehicle and a quadrotor show that Contextual Sampling achieves comparable or better tracking than Random Sampling with fewer trajectories, enabling real-time feasibility. Compared with Select-DPC~\cite{näf_choose_2025}, it achieves similar tracking accuracy at lower computational cost. In comparison with the full \acrshort{deepc} formulation without sampling, Contextual Sampling attains comparable tracking performance while requiring less computation, highlighting the benefit of efficient data selection in data-driven predictive control.
\end{abstract}


\section{Introduction}
\subsection{Motivation}
Real-time control of nonlinear robotic systems is critical for applications involving autonomous navigation, manipulation, and flight.
\acrfull{mpc} is widely used in robotic motion planning and control due to its ability to handle constraints and optimize performance \cite{camacho_model_2007}. However, deploying MPC in real-world robotic applications often requires an accurate system model, which may be unavailable or computationally expensive to maintain. This limitation is particularly critical for autonomous vehicle racing, where achieving precise trajectory tracking, high-speed stability, and dynamic adaptation requires accurate modeling of complex vehicle dynamics \cite{scheffe_sequential_2023, alrifaee_real-time_2018}. Model inaccuracies can lead to suboptimal lap times or instability, making real-time control a significant challenge.

To address the challenges related to model dependency, several data-driven control approaches have been proposed, offering alternatives to traditional model-based methods. These approaches leverage input-output data to learn system behavior, bypassing the need for explicit model identification. Among these approaches, \acrfull{deepc} \cite{coulson_data-enabled_2019} has garnered attention due to its ability to optimize system performance while respecting constraints. \acrshort{deepc} directly incorporates control objectives and system constraints into its formulation, making it particularly appealing for practical applications. However, its computational demands increase with the size and diversity of the dataset required to capture nonlinear behavior.

\subsection{Related Work}
In \cite{berberich_overview_2024}, Berberich and Allgöwer provide an overview of data-driven control with a focus on system-theoretic guarantees. They divide data-driven approaches for nonlinear systems into two classes: (1) approaches that exploit global linearity in higher-dimensional coordinates and (2) approaches that exploit local linearity.

The first class of approaches assume structural knowledge of nonlinearities or attempt to learn them. A data-driven framework extended to Hammerstein–Wiener systems by Berberich and Allgöwer \cite{berberich_trajectory-based_2020} shows that unknown nonlinearities can be approximated via kernel methods. Likewise, Rueda-Escobedo and Schiffer \cite{rueda-escobedo_data-driven_2020} present a data-driven internal model controller for second-order discrete Volterra systems. Both methods do not consider output nonlinearities. To address this limitation, Mishra et al. \cite{mishra_data-driven_2021} introduce a simulation approach for \acrfull{narx} systems, covering various nonlinear models, including Gaussian Processes and Neural Networks. Other approaches seek to represent system dynamics in higher-dimensional spaces. Lian et al. \cite{lian_nonlinear_2021} and Lian, Wang, and Jones \cite{lian_koopman_2021} employ reproducing kernel functions and the Koopman operator, respectively, enabling globally linear dynamics but at the cost of increased optimization complexity. Huang, Lygeros, and Dörfler \cite{huang_robust_2024} extend this idea by integrating regularized kernel methods with \acrshort{deepc} to capture nonlinearities via the representer theorem, resulting in a structured yet nonconvex optimization problem. Further extending \acrshort{deepc}, \cite{lazar_basis-functions_2024} incorporates general basis functions, proposing sparse regularization and ridge regression for computational efficiency. Lazar \cite{lazar_neural_2024} further advances this direction by using neural networks to learn implicit nonlinear bases, providing greater modeling flexibility. Collectively, these approaches extend \acrshort{deepc} by either embedding prior system knowledge through kernel methods and basis functions or enabling purely data-driven representations using neural networks.

The second class of approaches leverages the fact that nonlinear systems can be locally approximated by linear dynamics. 
Berberich et al.~\cite{berberich_linear_2022} extend Willems’ Fundamental Lemma to affine systems and propose an approach that updates data online for accurate local predictions. 
However, this requires persistently exciting input signals to ensure sufficient system exploration, and its real-time feasibility is not addressed. 
In parallel to our Contextual Sampling approach, Näf et al.~\cite{näf_choose_2025} have proposed a trajectory selection scheme that iteratively refines the selected data using an SQP-like procedure.
At each iteration, an open-loop prediction is used to update the selected subset before resolving the optimization problem. 
In contrast, our Contextual Sampling method focuses on real-time feasibility by selecting the most relevant trajectories in a single iteration based on both the current operating point and the future reference, achieving comparable tracking accuracy at lower computational cost.

\subsection{Contribution}
The main contributions of this work are summarized as follows:
\begin{itemize}
    \item We propose an efficient data selection strategy for \acrshort{deepc}, termed \emph{Contextual Sampling}, which dynamically selects the most relevant trajectories from a fixed dataset based on the system’s current state and reference. 
    \item The method operates on an adaptively chosen subset of data during each optimization, improving computational efficiency while preserving the representativeness of the full dataset. 
    \item We validate the approach on two robotic platforms—a scaled autonomous vehicle and a quadrotor—demonstrating real-time predictive control with high tracking accuracy. Contextual Sampling achieves comparable or better performance than Random Sampling, Select-DPC~\cite{näf_choose_2025}, and the full DeePC formulation without sampling, while requiring substantially less computation time.
\end{itemize}

\subsection{Structure}
The remainder of this paper is organized as follows. 
Section~II introduces the theoretical background of Data-Driven Predictive Control and details the proposed Contextual Sampling method. 
Section~III presents simulation studies on a scaled vehicle and a quadrotor to evaluate tracking accuracy and computational efficiency. 
Finally, Section~IV concludes the paper and discusses directions for future research.

\section{Methodology}\label{sec:methodology}

\paragraph*{Notation} 
We denote by \(\operatorname{col}(\cdot)\) the column-wise concatenation of vectors or matrices, 
\(\operatorname{colspan}(\cdot)\) the column space of a matrix, and \(\|\cdot\|_R\) the weighted 2-norm with weighting matrix \(R\). 
Furthermore, \(\mathbb{Z}_{>0}\) denotes the set of positive integers.

\subsection{Preliminaries}
\acrshort{deepc} is based on Behavioral Systems Theory \cite{markovsky_exact_2006} for discrete-time Linear Time-Invariant (LTI) systems whose input-output trajectories satisfy Willems' Fundamental Lemma \cite{willems_note_2005}, meaning that all valid trajectories can be reconstructed from a sufficiently rich set of past data.

\begin{definition}[Persistency of Excitation, \cite{coulson_data-enabled_2019, willems_note_2005}]
    Let \( L, T, m \in \mathbb{Z}_{>0} \) such that \( T \geq L \).  
    A signal \( u = \operatorname{col}(u_1, \dots, u_T) \in \mathbb{R}^{T m} \)  
    is \textit{persistently exciting} of order \( L \) if the Hankel matrix  
    \begin{equation}
        \mathscr{H}_L(u) :=
        \begin{bmatrix}
            u_1 & u_2 & \cdots & u_{T-L+1} \\
            u_2 & u_3 & \cdots & u_{T-L+2} \\
            \vdots & \vdots & \ddots & \vdots \\
            u_L & u_{L+1} & \cdots & u_T
        \end{bmatrix}
        \in \mathbb{R}^{L m \times (T-L+1)}
        \label{eq:hankel}
    \end{equation}
    is of full row rank.
\qed
\end{definition}

Here, \( T \) represents the length of the signal, \(L\) the length of each subtrajectory, and \( m \) the number of inputs. In other words, a persistently exciting signal excites the system such that the resulting input/output signal captures the whole system behavior. This condition bounds the length of the trajectory to \( T \geq (m+1)L-1 \). Subsequently, the result stated in Lemma \ref{lem:fundamental} (Fundamental Lemma) shows that the columns of the Hankel matrix span a subspace equal to the subspace of trajectories the underlying linear system can generate. We refer to the columns of the Hankel matrix as subtrajectories.

\begin{lemma} (Fundamental Lemma, \cite{willems_note_2005,coulson_data-enabled_2019}) \label{lem:fundamental}
Consider a controllable system $\mathscr{B}$. Let \( T, t \in \mathbb{Z}_{>0} \), and \( w = \operatorname{col}(u,y) \in \mathscr{B}_T \). Assume \( u \) to be \textit{persistently exciting} of order \( t+\bm{n}(\mathscr{B}) \). Then 
\[
\operatorname{colspan}(\mathscr{H}_t(w)) = \mathscr{B}_t.
\]
\end{lemma}

\( \mathscr{B}_T \) represents the set of trajectories truncated to length \( T \), and \( \bm{n}(\mathscr{B}) \) the order of the minimal input/output/state representation of the underlying linear system. Van Waarde et al. extend the Fundamental Lemma to mosaic-Hankel matrices \cite{van_waarde_willems_2020}, given that the input sequences are \textit{collectively persistently exciting} of order \( t+\bm{n}(\mathscr{B}) \), meaning the mosaic-Hankel matrix ${\mathscr{\tilde{H}}}_{L}$ has full row rank. A mosaic-Hankel matrix is a horizontal concatenation of multiple Hankel matrices. The latter extension is essential for our approach, as it allows us to select a subset of the original Hankel matrix. Additionally, Berberich et al. extend the Fundamental Lemma to the class of affine systems \cite{berberich_linear_2022}. The first extension allows us to build our data matrix by sampling from the Hankel matrix. The second extension is the basis for linearly approximating nonlinear systems locally.

To leverage \ref{lem:fundamental} (Fundamental Lemma), given a mosaic-Hankel matrix ${\mathscr{\tilde{H}}}_{L}$, we build the input and output Hankel matrices $\mathscr{\tilde{H}}_{L}(u), \mathscr{\tilde{H}}_{L}(y)$. These are then partitioned into past and future data of length \( L =T_\text{ini} + N \), respectively:

\begin{equation}
\begin{pmatrix}
\tilde{U}_\text{p} \\ \tilde{U}_\text{f}
\end{pmatrix}
:= \mathscr{\tilde{H}}_{T_{\text{ini}}+N}(u), \quad
\begin{pmatrix}
\tilde{Y}_\text{p} \\ \tilde{Y}_\text{f}
\end{pmatrix}
:= \mathscr{\tilde{H}}_{T_{\text{ini}}+N}(y).
\end{equation}

According to the Fundamental Lemma, the trajectory \( \operatorname{col}(u_\text{ini}, y_\text{ini}, u, y) \) belongs to the linear system \( \mathscr{B}_{T_\text{ini}+N} \) if and only if there exists \( g \) such that:

\begin{equation}
\begin{pmatrix}
\tilde{U}_{\text{p}} \\ \tilde{Y}_{\text{p}} \\ \tilde{U}_{\text{f}} \\ \tilde{Y}_{\text{f}}
\end{pmatrix}
g =
\begin{pmatrix}
u_{\text{ini}} \\ y_{\text{ini}} \\ u \\ y
\end{pmatrix}.
\label{eq:deepc-predictor}
\end{equation}

The signals \( u_\text{ini}, y_\text{ini} \) represent the last $T_\text{ini}$ initial inputs and outputs and \( u, y \) the future inputs and outputs. The first three blocks of equation \eqref{eq:deepc-predictor} can be solved for \( g \), given $u_\text{ini}, y_\text{ini}$, and $u$. The future outputs can be predicted by solving the last block of equation \eqref{eq:deepc-predictor} \( \tilde{Y}_\text{f}g = y \).
The Fundamental Lemma relies on a binary notion of persistency of excitation: a signal is either sufficiently exciting or not, depending on whether the associated Hankel matrix has full row rank~\cite{willems_note_2005}. 
However, in practical data-driven control applications, measurement noise and finite data lengths can lead to Hankel matrices that are nominally full rank but ill-conditioned, compromising numerical robustness. 
To address this, Coulson et~al.\ introduce the notion of \textit{quantitative persistency of excitation} that requires a strict lower bound on the smallest nonzero singular value of the data matrix, i.e.\ \(\sigma_{\min}(\mathscr{\tilde{H}}_{L}) \ge \alpha > 0\)~\cite{coulson_quantitative_2023}. 
This condition not only ensures rank sufficiency but also enforces a minimal level of excitation in every direction of the data subspace, effectively providing a signal-to-noise margin. 
In this work, we monitor the conditioning of the data matrices to qualitatively assess this requirement and discuss its implications for sampling in Section~\ref{sec:results}.

\subsection{Regularized \acrshort{deepc}}
In \acrshort{deepc}, this data-driven predictor is used as a nonparametric model. At each time step, optimization problem \eqref{eq:deepc-optimization-problem} is solved in a receding horizon fashion to compute the optimal inputs in terms of the given cost function. Our approach is based on the regularized formulation of \acrshort{deepc} from Coulson et al. \cite{coulson_data-enabled_2019}:

\begin{equation}
\begin{aligned}
\underset{g, u, y, \sigma_y}{\text{minimize}} \quad & 
\sum_{k=1}^{N} \|y_k - r_{k}\|_Q^2 + \sum_{k=0}^{N-1} \left( \|u_k\|_R^2 + \|\Delta u_k\|_R^2 \right) \\
& + \lambda_g \|g\|_2 + \lambda_\sigma \|\sigma_y\|_1, \\
\text{subject to} \quad & 
\begin{pmatrix}
\tilde{U}_{\text{p}} \\ \tilde{Y}_{\text{p}} \\ \tilde{U}_{\text{f}} \\ \tilde{Y}_{\text{f}} \\ \mymathbb{1}^\intercal
\end{pmatrix}
g = 
\begin{pmatrix}
u_{\text{ini}} \\ y_{\text{ini}} + \sigma_y \\ u \\ y \\ 1
\end{pmatrix} \\
& u_k \in \mathcal{U}, \quad \forall k \in \{0, \dots, N-1\}, \\
& y_k \in \mathcal{Y}, \quad \forall k \in \{1, \dots, N\}.
\end{aligned}
\label{eq:deepc-optimization-problem}
\end{equation}

The cost function consists of a tracking term to follow the reference \( r \), a penalty on \( u \) to minimize energy, a penalty on \( \Delta u \) for smooth input trajectories, and two regularization terms. We use the 2-norm instead of the 1-norm for regularization, as it is computationally more efficient \cite{schmitt_data_2023}. The regularization on \( g \) enables reliable performance for noisy data, while the cost on the slack variable $\sigma_y$ is intended as a penalty. The data-driven predictor is incorporated as an equality constraint describing the relation between \( u \) and \( y \). Note that a slack variable  $\sigma_y$ was added to ensure that the constraint is feasible in the case of nonlinear data. Last, constraints on \( u \) and \( y \) are encoded in the sets \( \mathcal{U} \) and \( \mathcal{Y} \), respectively.

\subsection{Contextual Sampling}
We propose the use of Contextual Sampling to address two issues. First, the number of subtrajectories in the Hankel matrix is equivalent to the length of \( g \). As a consequence, the number of optimization variables in the optimization problem increases with the size of the Hankel matrix, leading to greater computational complexity. Second, while approximating the system dynamics of a nonlinear system via equation \eqref{eq:deepc-predictor} works well for various applications \cite{coulson_data-enabled_2019, schmitt_data_2023}, there is still room for improvement, as shown by numerous works extending \acrshort{deepc} for nonlinear systems \cite{lazar_basis-functions_2024, lazar_neural_2024, huang_robust_2024, lian_nonlinear_2021}.

Contextual Sampling addresses these issues by selecting the data closest to the current initial trajectory and desired future reference, thereby reducing the amount of data used for prediction while staying close to the current and upcoming system dynamics. 
At each time step, we identify past trajectories that are simultaneously close to the current initial outputs and the future reference trajectory. 
Given the current initial outputs \( y_{\text{ini}} \) and the desired future reference \( r_{\text{f}} \), we define the distances to each subtrajectory \( (Y_{\text{p},i}, Y_{\text{f},i}) \) in the output Hankel matrices as:
\begin{equation}
\begin{aligned}
    d_{\text{p},i} &= \left\| Y_{\text{p},i} - y_{\text{ini}} \right\|_2, \\
    d_{\text{f},i} &= \left\| Y_{\text{f},i} - r \right\|_Q,
\end{aligned}
\end{equation}
where \( Q \) is the output weighting matrix used in the cost function of the controller. 
This ensures that output channels without a defined reference are automatically de-emphasized in the distance metric.
Both distance vectors are normalized to zero mean and unit variance, and combined into a single similarity measure:
\begin{equation}
    \tilde{d}_i = 
    \frac{d_{\text{p},i} - \mu_{\text{p}}}{\sigma_{\text{p}}} +
    \frac{d_{\text{f},i} - \mu_{\text{f}}}{\sigma_{\text{f}}},
\end{equation}
where \(\mu_{\text{p}}, \sigma_{\text{p}}\) and \(\mu_{\text{f}}, \sigma_{\text{f}}\) denote the means and standard deviations of the respective distance vectors.
We extract the \( N_\text{s} \) indices of the smallest entries of \( \tilde{d} \) and store them in the set \( \mathcal{I} \):
\begin{equation}
   \mathcal{I} = \underset{|\mathcal{I}| = n_\mathrm{s}}{\operatorname{argmin}} \;
   \sum_{i \in \mathcal{I}} \tilde{d}_i.
\end{equation}

The number of sampled subtrajectories \(n_\mathrm{s}\) is a tuning parameter. 
It should be chosen as small as possible while maintaining sufficient prediction accuracy. 
The softmin weighting introduces a small degree of randomness in the selection process, 
assigning higher probabilities to trajectories with smaller distances while still allowing 
exploration of nearby alternatives. 
This stochasticity promotes diversity among the selected trajectories. 
In Section~\ref{sec:results}, we will observe that choosing \( N_\text{s} \) creates a trade-off between computation time and tracking accuracy.
Using the selected subtrajectory indices \( \mathcal{I} = \{ i_1, i_2, \dots, i_{N_\text{s}} \} \), we construct the mosaic-Hankel matrices \( \tilde{U}_\text{p}, \tilde{Y}_\text{p}, \tilde{U}_\text{f}, \tilde{Y}_\text{f} \) from the original Hankel matrices \( U_\text{p}, Y_\text{p}, U_\text{f}, Y_\text{f} \):
\begin{equation}
    \begin{aligned}
        \tilde{U}_\text{p} &= U_{\text{p},:,\mathcal{I}}, &\tilde{Y}_\text{p} &= Y_{\text{p},:,\mathcal{I}}, \\
        \tilde{U}_\text{f} &= U_{\text{f},:,\mathcal{I}}, &\tilde{Y}_\text{f} &= Y_{\text{f},:,\mathcal{I}}.
    \end{aligned}
\end{equation}
This process is repeated at every time step to adapt the data to the current operating point and upcoming reference trajectory. 
We outline the combination of Contextual Sampling and \acrshort{deepc} in Algorithm~\ref{alg:contextual-sampling-deepc}.

\begin{algorithm}[tb]
\caption{\acrshort{deepc} with Contextual Sampling}
\label{alg:contextual-sampling-deepc}

\textbf{Input:} Hankel matrices $\{U_\text{p}, Y_\text{p}, U_\text{f}, Y_\text{f}\}$, reference trajectory $r_\text{f}$, past input/output data $\{u_\text{ini}, y_\text{ini}\}$ of length $T_\text{ini}$, feasible sets $\mathcal{U}$ and $\mathcal{Y}$, cost matrices $Q$ and $R$, number of samples $N_\text{s}$.

\begin{algorithmic}[1]
\STATE \textbf{Compute distance vectors:}
\[
\begin{aligned}
    d_{\text{p},i} &= \left\| Y_{\text{p},i} - y_{\text{ini}} \right\|_2, \\
    d_{\text{f},i} &= \left\| Y_{\text{f},i} - r \right\|_Q
\end{aligned}
\]
\STATE \textbf{Normalize and combine:}
\[
\tilde{d}_i =
\frac{d_{\text{p},i} - \mu_{\text{p}}}{\sigma_{\text{p}}} +
\frac{d_{\text{f},i} - \mu_{\text{f}}}{\sigma_{\text{f}}}
\]
\STATE \textbf{Extract closest trajectories:}
\[
\mathcal{I} = \underset{|\mathcal{I}| = N_\text{s}}{\operatorname{argmin}} \;
\sum_{i \in \mathcal{I}} \tilde{d}_i
\]
\STATE \textbf{Construct mosaic-Hankel matrices:}
\[
\begin{aligned}
\tilde{U}_\text{p} &= U_{\text{p},:,\mathcal{I}}, \quad &\tilde{Y}_\text{p} &= Y_{\text{p},:,\mathcal{I}}, \\
\tilde{U}_\text{f} &= U_{\text{f},:,\mathcal{I}}, \quad &\tilde{Y}_\text{f} &= Y_{\text{f},:,\mathcal{I}}
\end{aligned}
\]
\STATE \textbf{Solve \eqref{eq:deepc-optimization-problem}} for $g^\star$.
\STATE \textbf{Compute the optimal input sequence:}
\[
u^\star = \tilde{U}_\text{f} g^\star
\]
\STATE \textbf{Apply next input} $u(t)$.
\STATE \textbf{Update time step:} $t \gets t + 1$, and update $u_\text{ini}$ and $y_\text{ini}$ with the most recent $T_\text{ini}$ measurements.
\STATE \textbf{Repeat from line 1.}
\end{algorithmic}
\end{algorithm}

Our proposed method builds on \acrshort{deepc} to apply data-driven control to nonlinear systems. By selecting past trajectories closest to the current operating point we implicitly leverage the idea that, locally, the nonlinear system behaves approximately linearly. This aligns with the theoretical framework of \cite{berberich_linear_2022}, which extends the Fundamental Lemma to affine systems and demonstrates that, under suitable assumptions, the input-output behavior of a nonlinear system can be well-approximated by its local linearization. Moreover, their results establish practical stability guarantees for data-driven \acrshort{mpc} applied to nonlinear systems, ensuring that the closed-loop system remains near the optimal reachable equilibrium.

While our approach shares this fundamental insight, Contextual Sampling dynamically selects relevant data rather than modifying the underlying data-driven model formulation. This raises theoretical questions about the impact of data selection on stability and robustness, particularly how the selected subset influences prediction error and long-term closed-loop performance. Our future work will build on the results of \cite{berberich_linear_2022} to investigate theoretical guarantees for Contextual Sampling.

\subsection{Implementation and Preprocessing}
We introduce several modifications to improve numerical stability and 
practical applicability of \acrshort{deepc} with Contextual Sampling.

\paragraph*{Normalization}
We normalize all input and output signals using z-score normalization, 
computed once during dataset initialization. 
The mean and standard deviation obtained from the training data are reused 
to normalize the initial input-output sequences and the reference trajectories 
throughout the experiments.

\paragraph*{Trajectory Alignment}
Before the experiments, we preprocess all recorded trajectories to ensure spatial consistency. 
We transform each trajectory into a local reference frame such that the final pose of its initial output trajectory aligns with a common origin. 
In the vehicle experiments, we shift the final pose to \((x,\,y,\,\psi) = (0,\,0,\,0)\); 
in the quadrotor experiments, we align the final position to \((x,\,y,\,z) = (0,\,0,\,1)\).  
This alignment enables Contextual Sampling to compare trajectories directly using Euclidean distance, ensuring that similarity is evaluated consistently across experiments.

\paragraph*{Incremental Input Representation}
To express the system in control-affine form and improve local linearity, 
the inputs are reformulated as increments relative to the previous control 
action. 
The transformed outputs include both the original measured outputs and the 
current inputs concatenated, preserving the input influence within the 
data-driven model.

\paragraph*{Regularization Scaling}
We scale the regularization parameter \(\lambda_g\) by the number of selected 
subtrajectories \(n_\mathrm{s}\), i.e.,
\[
\lambda_g = \bar{\lambda}_g\,n_\mathrm{s},
\]
where \(\bar{\lambda}_g\) is a constant tuning factor. 
This scaling maintains consistent weighting of the regularization term across 
different dataset sizes, avoiding the need for retuning when varying the number 
of trajectories.

\section{Case Study: Trajectory Tracking for Vehicles and Quadrotors}

\subsection{Experimental Setup}
We evaluate \acrshort{deepc} with Contextual Sampling and Random Sampling, and Select-DPC across two robotic platforms—a scaled vehicle and a quadrotor. Each configuration is simulated \num{10} times with different random seeds shared across both sampling methods for consistent statistical comparison. All optimization problems are solved using OSQP~\cite{stellato_osqp_2020}.

\paragraph*{General Setup}
Both systems use the regularized \acrshort{deepc} formulation presented in Section~\ref{sec:methodology}. 
Table~\ref{tab:params} summarizes the controller parameters and constraints used in the vehicle and quadrotor experiments. 
Each dataset contains \SI{300}{\second} of input–output data. 
The number of sampled subtrajectories \(n_\mathrm{s}\) is varied to analyze the trade-off between dataset size, prediction accuracy, and computational efficiency. 
Starting from a small dataset, we gradually decreased \(n_\mathrm{s}\) until all methods failed—manifested as infeasibility for the vehicle and ground contact for the quadrotor—and then increased it in steps of 30. Accordingly, we test \(n_\mathrm{s} \in \{30, 60, 90\}\) for the vehicle and \(n_\mathrm{s} \in \{40, 70, 100\}\) for the quadrotor. 
For both case studies, performance improvements flattened out after two such increases, indicating an effective balance between computational cost and prediction accuracy. 
The regularization parameters \(\lambda_g\) and \(\lambda_\sigma\) are tuned separately for each sampling method and value of \(n_\mathrm{s}\) via a grid search over the range \([10^{-3}, 10^{3}]\). 
Because the optimal values differ between configurations, the table lists only parameters common to all experiments.
\begin{table}[t]
    \centering
    \caption{Controller parameters and constraints used in the case studies (excluding regularization parameters).}
    \label{tab:params}
    \begin{tabular}{lcc}
        \toprule
        \textbf{Parameter} & \textbf{Vehicle} & \textbf{Quadrotor} \\
        \midrule
        \(T_s\) & \SI{0.1}{\second} & \SI{0.04}{\second} \\
        \(T_{\mathrm{ini}},\, N\) & \(5,\,10\) & \(5,\,15\) \\
        \(Q\) & \(\operatorname{diag}(1,\,1,\,0.1,\,0.1)\) & \(\operatorname{diag}(1,\,1,\,1)\) \\
        \(R\) & \(\operatorname{diag}(0.1,\,1)\) & \(\operatorname{diag}(0.1,\,0.1,\,0.1,\,0)\) \\
        \bottomrule
    \end{tabular}
\end{table}
\paragraph*{Vehicle Simulation}
We simulate the vehicle in the open-source \textit{RoboRacer} formerly known as \textit{F1TENTH gym environment}~\cite{okelly_f1tenth_2019}. 
The dynamics follow the single-track model~\cite{althoff_commonroad_2017} 
with state vector \((x, y, v, \psi)\), representing the vehicle's position, velocity, 
and yaw angle, and control inputs \((a, \delta)\) denoting acceleration 
and steering angle, respectively. The state vector also serves as our output vector.
The task is to follow the São Paulo raceline~\cite{betz_autonomous_2022}. 
To collect training data, the vehicle is simulated for \SI{300}{\second} using randomized acceleration and steering commands to promote persistency of excitation. 
To assess robustness against measurement noise, we add artificial measurement noise with standard deviations of \(\sigma_{x,y} = \SI{0.05}{\meter}\), \(\sigma_{v} = \SI{0.05}{\meter\per\second}\), and \(\sigma_{\psi} = \SI{0.6}{\degree}\) to the position, velocity, and orientation measurements at each control step, and randomize the initial pose slightly.

\paragraph*{Quadrotor Simulation}
The quadrotor experiments are performed in the \textit{gym-pybullet-drones} environment~\cite{panerati_learning_2021} using the \textit{Crazyflie~2.x} model with six-degree-of-freedom rigid-body dynamics. 
The quadrotor dynamics are modeled with the state vector
\((x, y, z, v_x, v_y, v_z, \phi, \theta, \psi, \omega_x, \omega_y, \omega_z)\), 
representing position, linear velocity, orientation (roll, pitch, yaw), 
and angular velocity, respectively. 
The control inputs are the collective thrust and body torques. 
Only the position components \((x, y, z)\) are used as measured outputs. 
A built-in PID controller stabilizes attitude and thrust, while \acrshort{deepc} provides reference thrust and orientation commands for trajectory tracking. 
Training data are obtained from \SI{300}{\second} of flight with randomized waypoint sequences to excite relevant motion. 
During evaluation, the drone follows a smooth three-dimensional figure-eight trajectory of radius \(\SI{0.3}{\meter}\). 
To test robustness against dynamics noise, we apply wind disturbances as random external forces generated at each control step, drawn from a zero-mean normal distribution with a standard deviation of \(\sigma_{\mathrm{wind}} = \SI{0.01}{\newton}\).

\begin{figure}[t]
  \centering
  \begin{subfigure}[t]{0.48\linewidth}
    \centering
    \includegraphics[width=\linewidth]{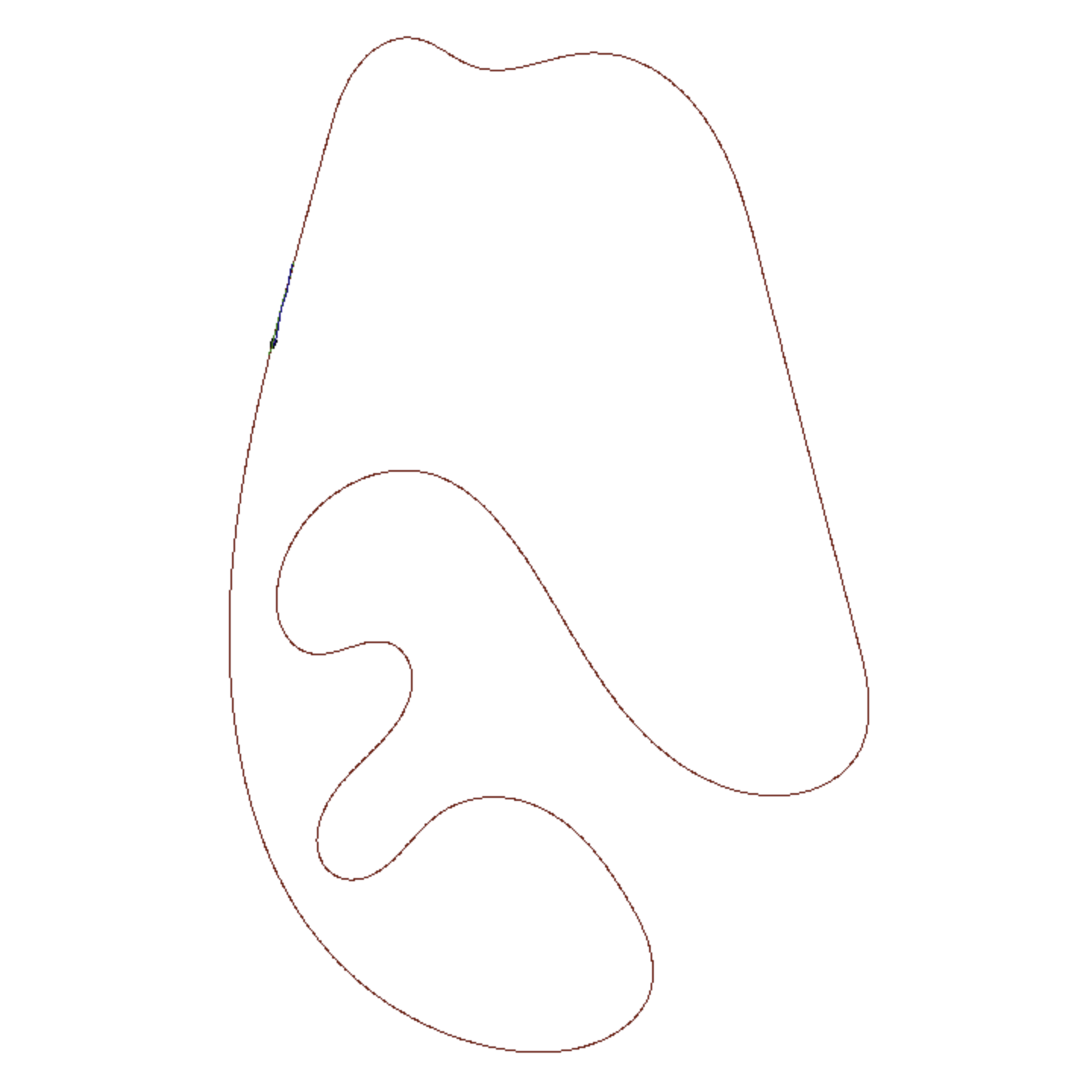}
    \caption{Vehicle reference trajectory (São Paulo track).}
    \label{fig:vehicle_ref}
  \end{subfigure}\hfill
  \begin{subfigure}[t]{0.48\linewidth}
    \centering
    \includegraphics[width=\linewidth]{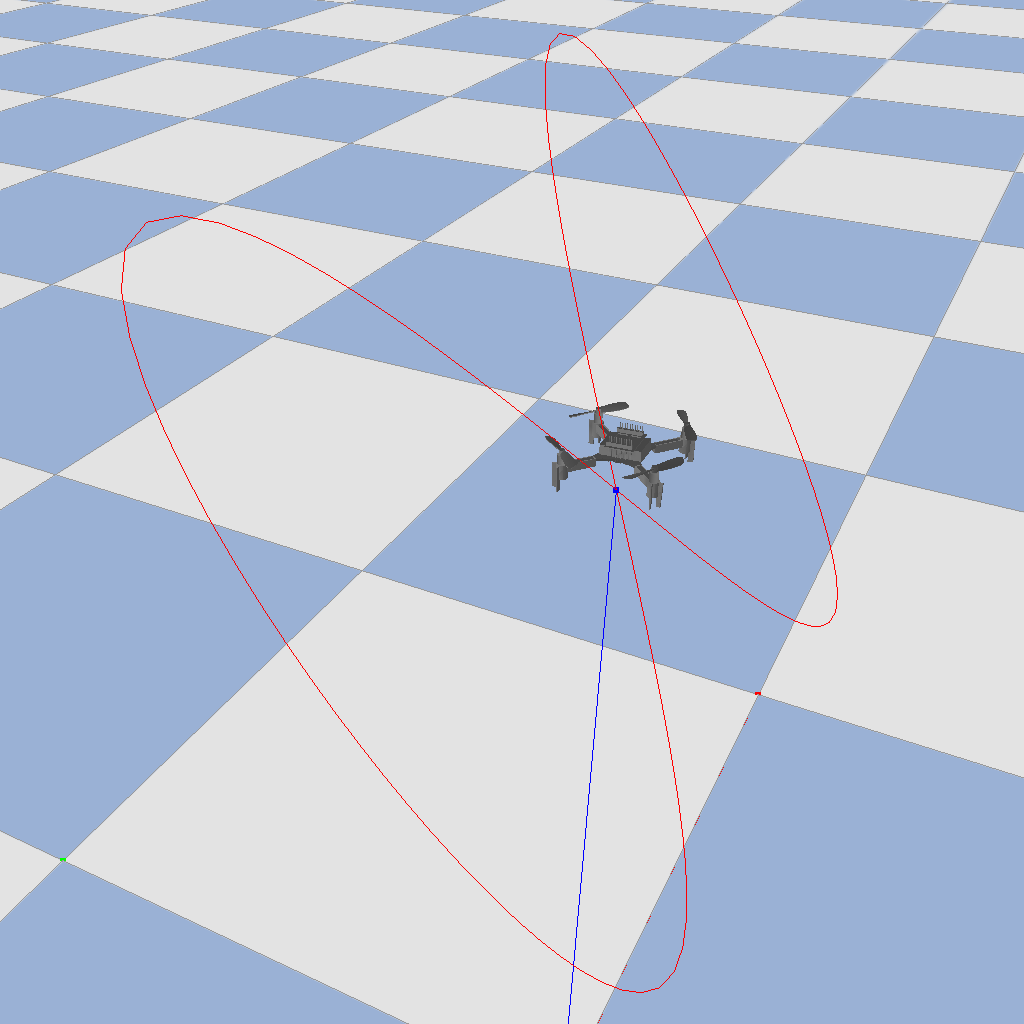}
    \caption{Quadrotor reference trajectory (3D figure-eight).}
    \label{fig:drone_ref}
  \end{subfigure}
  \caption{Reference trajectories used in the case studies for the vehicle and quadrotor experiments.}
  \label{fig:reference_trajectories}
\end{figure}

Fig.~\ref{fig:reference_trajectories} illustrates the reference paths used for the vehicle and quadrotor experiments. 
The vehicle follows the two-dimensional São Paulo raceline, while the quadrotor tracks a smooth three-dimensional figure-eight path.

\subsection{Results} \label{sec:results}
\begin{figure*}[t]
    \centering
    \begin{subfigure}{0.48\linewidth}
        \centering
        \includegraphics[width=\linewidth]{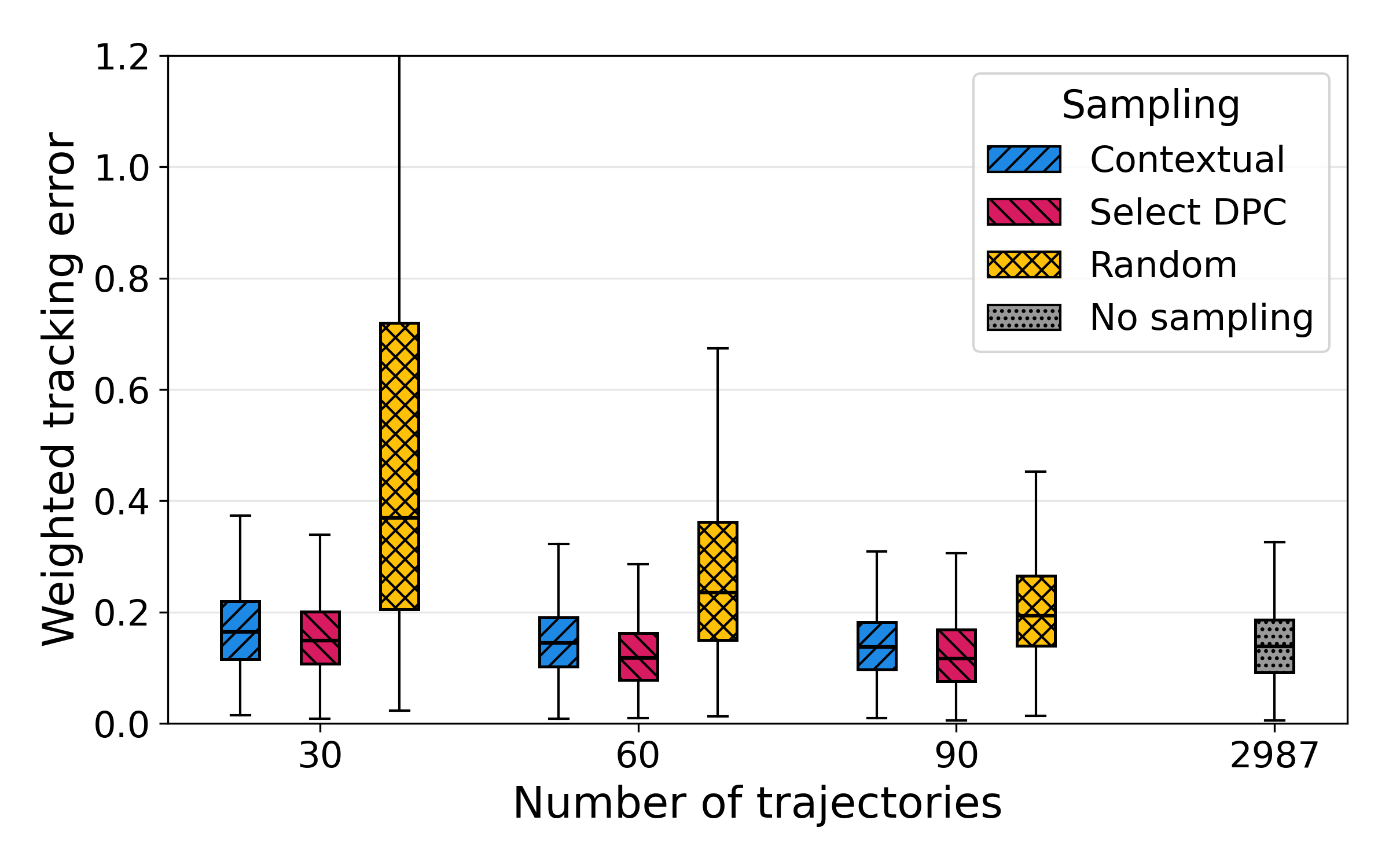}
        \caption{Vehicle}
        \label{fig:vehicle_errors_by_numtraj}
    \end{subfigure}
    \hfill
    \begin{subfigure}{0.48\linewidth}
        \centering
        \includegraphics[width=\linewidth]{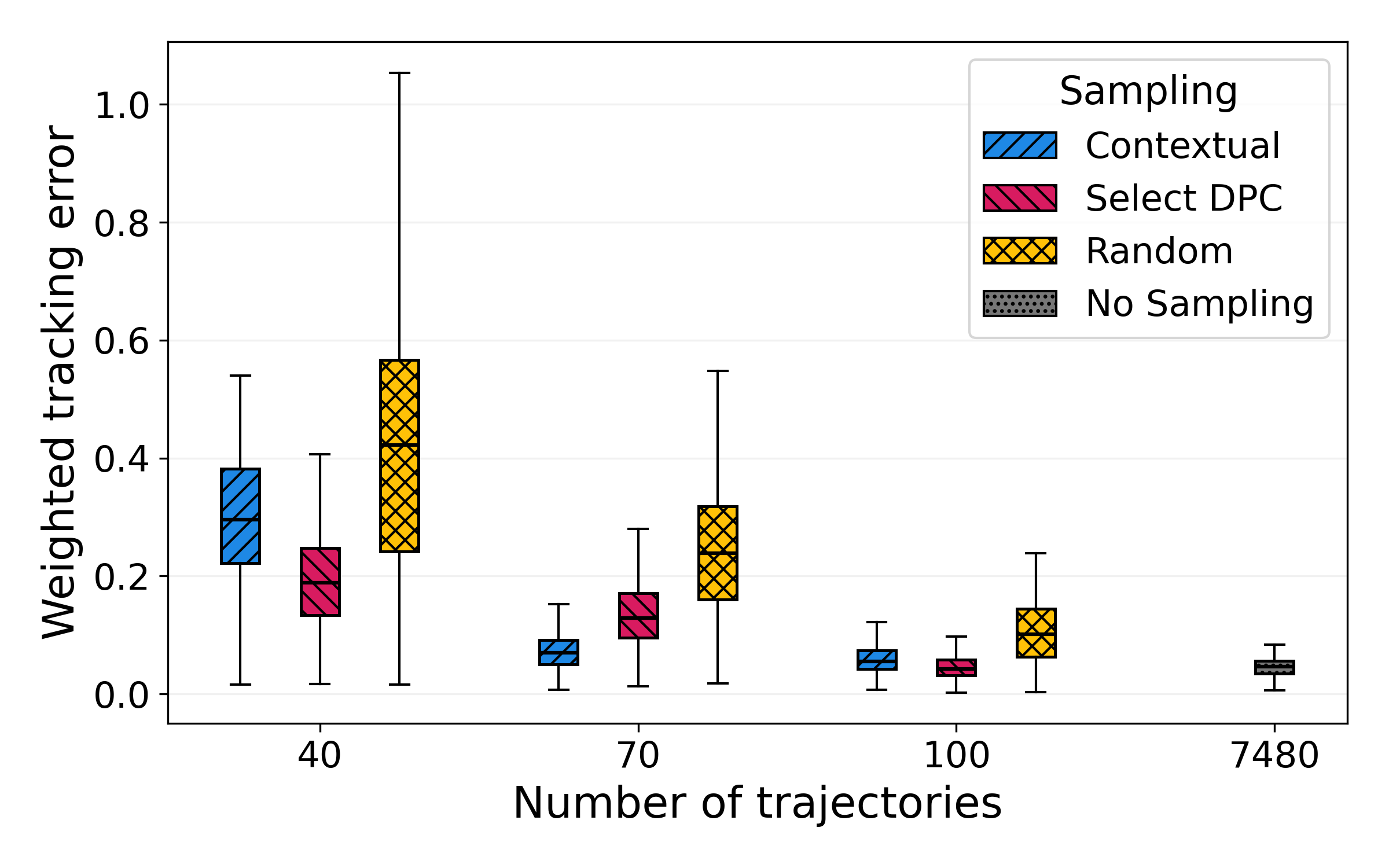}
        \caption{Quadrotor}
        \label{fig:drone_errors_by_numtraj}
    \end{subfigure}
    \caption{Closed-loop tracking error distributions across different numbers of trajectories and sampling methods for (a) the vehicle and (b) the quadrotor.}
    \label{fig:tracking_errors_by_numtraj}
\end{figure*}

We evaluate the closed-loop tracking performance of \acrshort{deepc} with Contextual Sampling, Select-DPC, and Random Sampling. 
For each method and number of trajectories, we perform ten independent simulations with different random seeds. 
At each time step \(t\), the weighted tracking error is computed as
\begin{equation}
    e_t = \| y_t - r_t \|_Q,
    \label{eq:tracking-error}
\end{equation}
where \(y_t\) and \(r_t\) denote the measured and reference outputs, respectively, and \(Q\) is the same weighting matrix as used in the cost function in equation \eqref{eq:deepc-optimization-problem}. 
For every configuration, the instantaneous errors from all time steps and simulation runs are concatenated, and their distributions are visualized as boxplots showing the median, quartiles, and overall variability (see Fig.~\ref{fig:vehicle_errors_by_numtraj} and~\ref{fig:drone_errors_by_numtraj}).
The boxplots in Fig.~\ref{fig:tracking_errors_by_numtraj} display the distribution of cumulative tracking errors aggregated over all time steps and simulation runs.

\begin{table*}[htb]
  \centering
  \caption{Per-step computation time percentiles for the vehicle and quadrotor (milliseconds).}
  \label{tab:solve_times_vehicle_quadrotor}

  \begin{subtable}[t]{0.46\textwidth}
    \centering
    \caption{Vehicle}
    \label{tab:solve_times_vehicle}
    \setlength{\tabcolsep}{4pt} 
  \begin{tabular}{r S[table-format=3.2] S[table-format=3.2] S[table-format=3.2] S[table-format=3.2] S[table-format=3.2] S[table-format=3.2]}
    \toprule
    & \multicolumn{2}{c}{Contextual} & \multicolumn{2}{c}{Select DPC} & \multicolumn{2}{c}{Random} \\
    \cmidrule(lr){2-3}\cmidrule(lr){4-5}\cmidrule(lr){6-7}
    {$N_{\mathrm{s}}$} & {p99 [ms]} & {Max [ms]} & {p99 [ms]} & {Max [ms]} & {p99 [ms]} & {Max [ms]} \\
    \midrule
    30 &  3.12 & 46.96 & 15.72 & 63.71 &  5.05 & 39.12 \\
    60 &  4.18 & 79.83 & 80.03 & 182.96 &  4.85 & 99.69 \\
    90 &  6.65 & 135.35 & 64.81 & 138.53 &  5.64 & 133.54 \\
    \bottomrule
  \end{tabular}
  \end{subtable}
  \hspace{0.06\textwidth} 
  \begin{subtable}[t]{0.46\textwidth}
    \centering
    \caption{Quadrotor}
    \label{tab:solve_times_quadrotor}
    \setlength{\tabcolsep}{4pt}
    \begin{tabular}{r S[table-format=3.2] S[table-format=3.2] S[table-format=3.2] S[table-format=3.2] S[table-format=3.2] S[table-format=3.2]}
    \toprule
    & \multicolumn{2}{c}{Contextual} & \multicolumn{2}{c}{Select DPC} & \multicolumn{2}{c}{Random} \\
    \cmidrule(lr){2-3}\cmidrule(lr){4-5}\cmidrule(lr){6-7}
    {$N_{\mathrm{s}}$} & {p99 [ms]} & {Max [ms]} & {p99 [ms]} & {Max [ms]} & {p99 [ms]} & {Max [ms]} \\
    \midrule
    40 & 34.96 & 101.85 & 44.47 & 74.94 & 31.93 & 93.50 \\
    70 & 72.51 & 204.90 & 115.02 & 142.55 & 62.88 & 264.04 \\
    100 & 102.86 & 300.65 & 159.05 & 177.75 & 98.29 & 321.75 \\
    \bottomrule
  \end{tabular}
  \end{subtable}
\end{table*}

For both systems, the closed-loop tracking performance exhibits the same qualitative trend. 
As the number of sampled subtrajectories \(n_\mathrm{s}\) increases, the tracking error initially decreases before reaching a plateau. 
At small \(n_\mathrm{s}\), the controller operates on a highly reduced dataset, which can lead to poor local approximations of the system dynamics and consequently higher tracking errors. 
As \(n_\mathrm{s}\) increases, both Contextual Sampling and Select-DPC outperform Random Sampling, achieving lower median errors and smaller variance across runs. 
This demonstrates that selecting trajectories based on their relevance to the current operating point improves predictive consistency and closed-loop behavior.

For the vehicle (Fig.~\ref{fig:vehicle_errors_by_numtraj}), performance improves substantially from \(n_\mathrm{s}=30\) to \(n_\mathrm{s}=60\), while further increasing to \(n_\mathrm{s}=90\) yields only marginal gains. 
Among the sampling strategies, Select-DPC achieves the lowest median tracking error in all configurations, followed closely by Contextual Sampling, whereas Random Sampling exhibits considerably higher variance, particularly for small \(n_\mathrm{s}\). 
Both informed sampling methods achieve comparable or better accuracy than using all available trajectories without sampling, while requiring significantly less computation time.

For the quadrotor (Fig.~\ref{fig:drone_errors_by_numtraj}), a similar trend is observed, with tracking accuracy improving from \(n_\mathrm{s}=40\) to \(n_\mathrm{s}=70\) and saturating at \(n_\mathrm{s}=100\). 
Select-DPC slightly outperforms Contextual Sampling in most configurations, except for \(n_\mathrm{s}=70\), where Contextual Sampling achieves a lower median error. 
Since Select-DPC again performs best at \(n_\mathrm{s}=100\), this deviation is likely due to imperfect parameter tuning during the grid search.
Across both systems, both informed sampling strategies outperform Random Sampling and approach or even outperform the performance of the full DeePC without sampling, demonstrating that dynamic data selection can maintain prediction accuracy while substantially reducing computational cost.

The effect of the number of subtrajectories on computation time is summarized in Table~\ref{tab:solve_times_vehicle_quadrotor}. 
All computation times were measured on a laptop equipped with an Apple M3 Pro CPU and \SI{36}{\giga\byte} of unified memory running macOS~15. 
To assess real-time capability, we consider the 99th-percentile (\(p_{99}\)) per-step computation time, as it reflects the worst-case performance under normal operating conditions while remaining robust to rare outliers that can distort the maximum value (\(p_{100}\)). 

For the vehicle, \(p_{99}\) remains below \SI{10}{\milli\second} for all configurations, well within the control-loop period of \SI{100}{\milli\second}. 
Computation time increases moderately with the number of subtrajectories, and all sampling strategies show near-linear scaling with \(n_\mathrm{s}\). 
Contextual Sampling achieves the lowest solve times, followed by Random Sampling, while Select-DPC is slower—particularly at \(n_\mathrm{s}=60\)—due to the additional optimization required in its selection process.

For the quadrotor, absolute computation times are higher and exceed the \SI{40}{\milli\second} bound for all but the smallest configuration. 
At \(n_\mathrm{s}=40\), \(p_{99}\) ranges from \SI{31.9}{\milli\second} (Random) to \SI{44.5}{\milli\second} (Select-DPC), and grows roughly linearly with \(n_\mathrm{s}\), reaching up to \SI{159}{\milli\second} for Select-DPC at \(n_\mathrm{s}=100\). 
Across all configurations, Select-DPC exhibits the highest computation times, while Contextual and Random Sampling remain similar and lower. 
Overall, computation time scales predictably with \(n_\mathrm{s}\), and all methods remain feasible for control rates typical of these systems.

\paragraph*{Observations on quantitative persistency of excitation}
We examined the minimum singular values \(\sigma_{\min}\) of the normalized mosaic Hankel matrices for Random and Contextual Sampling. 
For the vehicle, a heuristic lower bound \(\tilde\sigma_{\min,\mathrm{thresh}}\) was estimated from measurement noise and normalization scales. 
For the quadrotor, wind disturbances enter through the dynamics, so an analogous bound in output space would require propagating the force variance through a local linearization; we therefore focus on qualitative trends. 
Across both platforms, \(\sigma_{\min}\) does not differ systematically between sampling methods for a fixed \(N_{\mathrm s}\), yet Contextual Sampling achieves comparable or better tracking with fewer trajectories—thus operating at higher \(\sigma_{\min}\) and lower computational cost. 

It should be noted that for nonlinear systems, quantitative persistency of excitation can only be interpreted as a heuristic measure of data richness and numerical conditioning rather than a strict theoretical requirement. 
In practice, good conditioning depends strongly on the collected data: if trajectories are sufficiently dense but not redundant, conditioning remains adequate even for smaller datasets, and this can be further improved through preprocessing to remove near-duplicate trajectories. 
Poor conditioning can otherwise be mitigated by incorporating a backup dataset that is persistently exciting or by increasing the number of trajectories until the data matrix becomes sufficiently well-conditioned.

\subsection{Discussion}
While our results highlight the advantages of Contextual Sampling in improving computational efficiency and tracking accuracy compared to Random Sampling, Select-DPC achieves similar and sometimes even better performance. 
This is expected, as its optimization-based selection provides highly relevant data subsets. 
However, this comes at the cost of increased computation time, particularly for larger datasets. 
Therefore, when real-time feasibility is critical, Contextual Sampling offers a favorable trade-off between performance and computational efficiency.

\section{Conclusion}
This work introduced \emph{Contextual Sampling}, a data-driven trajectory selection strategy that complements \acrshort{deepc} by improving its computational efficiency for nonlinear robotic systems.
By dynamically selecting the most relevant subtrajectories based on the current operating point and future reference, the proposed approach reduces computational cost while maintaining or improving tracking accuracy compared to \acrshort{deepc} without sampling. 
Through simulation studies on both a scaled vehicle and a quadrotor, we demonstrated that Contextual Sampling achieves comparable or superior performance to Random Sampling with notably fewer trajectories, thereby improving real-time feasibility. 
A comparison with Select-DPC \cite{näf_choose_2025} revealed comparable tracking performance across both methods. Select-DPC achieved higher tracking accuracy in most configurations but required greater computational effort.
Moreover, we qualitatively examined the conditioning of the data matrices in relation to the quantitative notion of persistency of excitation, showing that Contextual Sampling operates effectively in regimes associated with higher minimum singular values. 
Future work will investigate the theoretical implications of dynamic data selection on stability and robustness, and further explore extensions to higher-dimensional systems and real-world robotic platforms.




\bibliographystyle{IEEEtran}
\bibliography{ref}

\end{document}